\documentclass[sigconf]{acmart}

\AtBeginDocument{%
  }

\setcopyright{acmlicensed}
\copyrightyear{2018}
\acmYear{2018}
\acmDOI{XXXXXXX.XXXXXXX}

\acmConference[Conference acronym 'XX]{Make sure to enter the correct
  conference title from your rights confirmation emai}{June 03--05,
  2018}{Woodstock, NY}
\acmISBN{978-1-4503-XXXX-X/18/06}

\usepackage[ruled]{algorithm2e}
\usepackage{subfigure}
\usepackage{booktabs}
\usepackage{graphicx}

\newcommand{\ie}{\emph{i.e.},\xspace}

\SetKwProg{Fn}{Function}{}{end}
\SetKwFunction{CHIPANNS}{C-HiPANNS}
\SetKwFunction{QHIPANNS}{Q-HiPANNS}

\setcopyright{none}
\renewcommand\footnotetextcopyrightpermission[1]{}
\begin{document}

\title{GPU-accelerated Multi-relational Parallel Graph Retrieval for Web-scale Recommendations}


\author{Zhuoning Guo$^1$, Guangxing Chen$^2$, Qian Gao$^2$, Xiaochao Liao$^2$, Jianjia Zheng$^2$, Lu Shen$^2$, Hao Liu$^{1,3}$}
\affiliation{%
    \institution{
    $^1$The Hong Kong University of Science and Technology (Guangzhou),
    $^2$Baidu Inc.,
    }
    \institution{
    $^3$The Hong Kong University of Science and Technology
    }
    \country{}
}
\email{zguo772@connect.hkust-gz.edu.cn,}
\email{{chenguangxing,gaoqian05,liaoxiaochao,zhengjianjia,shenlu}@baidu.com,liuh@ust.hk}








\renewcommand{\shortauthors}{}

\begin{abstract}

Web recommendations provide personalized items from massive catalogs for users, which rely heavily on retrieval stages to trade off the effectiveness and efficiency of selecting a small relevant set from billion-scale candidates in online digital platforms.
As one of the largest Chinese search engine and news feed providers, Baidu resorts to Deep Neural Network (DNN) and graph-based Approximate Nearest Neighbor Search (ANNS) algorithms for accurate relevance estimation and efficient search for relevant items.
However, current retrieval at Baidu fails in comprehensive user-item relational understanding due to dissected interaction modeling, and performs inefficiently in large-scale graph-based ANNS because of suboptimal traversal navigation and the GPU computational bottleneck under high concurrency.
To this end, we propose a GPU-accelerated Multi-relational Parallel Graph Retrieval (GMP-GR) framework to achieve effective yet efficient retrieval in web-scale recommendations.
First, we propose a multi-relational user-item relevance metric learning method that unifies diverse user behaviors through multi-objective optimization and employs a self-covariant loss to enhance pathfinding performance.
Second, we develop a hierarchical parallel graph-based ANNS to boost graph retrieval throughput, which conducts breadth-depth-balanced searches on a large-scale item graph and cost-effectively handles irregular neural computation via adaptive aggregation on GPUs.
In addition, we integrate system optimization strategies in the deployment of GMP-GR in Baidu.
Extensive experiments demonstrate the superiority of GMP-GR in retrieval accuracy and efficiency.
Deployed across more than twenty applications at Baidu, GMP-GR serves hundreds of millions of users with a throughput exceeding one hundred million requests per second.

\end{abstract}



\maketitle

\section{Introduction}

Recently, recommender systems have become essential tools for helping users navigate various options and discover relevant products or content~\cite{wu2022survey}.
However, large-scale relevance evaluation based on user preference is computationally prohibitive in real-world recommendation applications~\cite{zhu2018learning}.
For timely answering requirements, computing relevance values of billions of items is unacceptable~\cite{liu2021jizhi}.
To balance effectiveness and efficiency, retrieval-and-ranking recommendation frameworks~\cite{cheng2016wide,ma2020off} have gained widespread adoption in modern industrial systems, such as YouTube~\cite{covington2016deep}, LinkedIn~\cite{borisyuk2016casmos}, and Pinterest~\cite{eksombatchai2018pixie}.
The first stage~(\ie retrieval) focuses on retrieving a thousand-scale relevant subset from the entire billion-scale item catalog, which then serves as the candidates for further ordering in the second stage~(\ie ranking).
In particular, retrieval plays an important role in real-time and satisfactory answers for large-scale queries.
Effective retrieval can significantly simplify the search space for potential answers while maintaining personalization and serendipity, thus reducing the ranking difficulty in producing reliable final recommendations.

\begin{figure*}
\setlength{\abovecaptionskip}{0.1cm}
\setlength{\belowcaptionskip}{-0.3cm}
    \centering
    \includegraphics[width=\linewidth]{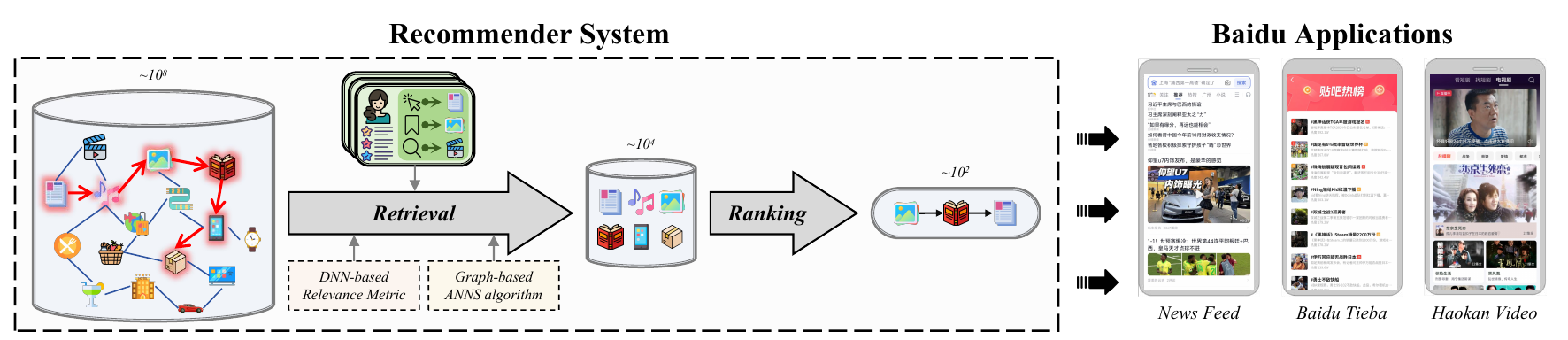}
    \caption{Applications of recommender systems at Baidu.}
    \label{fig:application}
\end{figure*}

In this work, we focus on the retrieval stage in Baidu's web-scale recommendation scenarios, e.g., news feed, Tieba, and video clips. Recent works~\cite{tan2020fast,chen2022approximate} establish the retrieval framework in a training-and-search manner.
Based on the complex relational dependencies modeling ability of Deep Neural Network~(DNN)~\cite{zhang2019deep}, they attempt to train a DNN-based metric for more accurate user-item relevance estimation compared with simple functions~\cite{dehghani2017neural,tay2018latent}.
Nevertheless, real-time retrieval is difficult due to complicated neural computations on massive candidates.
To solve the problem, researchers have been exploring the potential of Approximate Nearest Neighbor Search~(ANNS) algorithms to mitigate extensive computations, which search for relatively relevant items for each query user by selectively visiting instead of the entire item base.
Early works propose tree-based ANNS to organize data points in a tree-like structure by recursively splitting the dimensional space~\cite{friedman1977algorithm}.
However, they underperform in web-scale recommendations that aim to retrieve divergent and high-dimensional information. Because they are sensitive to complex feature space and item relationships~\cite{wang2021comprehensive}.
To improve retrieval effectiveness, the latest frameworks leverage graph-based ANNS algorithms that build similarity-based candidate graphs to efficiently expand the candidate set via graph traversal.
For example, HNSW~\cite{malkov2018efficient} hierarchically establishes a multi-layer indexing graph, grounding coarse-to-fine navigation towards the most satisfactory candidates.

However, we confront three challenges in developing an accurate and efficient retrieval framework for web-scale recommendations.
\textit{(1)~Unified modeling of multi-relational user-item relevance.} Baidu recommendation applications consist of dozens of distinct and related behaviors~(e.g., click, browse) that are typically modeled by a tailored metric for estimating relevance between users and items. However, the independent modeling of these user-item interactions fails to capture the semantic relations of diverse user behaviors, which are mutually beneficial between them~\cite{tan2020fast}. Besides, training multiple behavior-specific metrics incurs exponential efforts in system updates. Thus, the first challenge lies in cost-effectively multi-relational metric joint training for more accurate relevance estimation.
\textit{(2)~Efficient search on large-scale graph item base.} Baidu devises Graph-based ANNS to traverse relevant candidate items from a graph item base. Practically, a trade-off problem arises that increasing the online retrieval throughput will restrict the search efforts, \ie the graph traversal steps. However, existing methods make delivering satisfactory items in limited steps difficult, especially for billion-scale graphs in web-scale recommender systems.
On the one hand, neural relevance metrics often violate the triangle inequality, a property that ensures optimal pathfinding in graph traversal~\cite{chen2022approximate}. This violation forces the exploration of redundant edges, decreasing the possibility of reaching distant yet relevant candidates.
On the other hand, graph-based ANNS usually search for relevant items greedily, which may be stuck in a narrow space after a short-length traversal among a large-scale graph.
The inability to balance search depth and breadth degrades the quality of real-time recommendations. Addressing this challenge requires developing a search-compatible relevance metric and a deep-and-broad traversal algorithm for efficient graph search.
\textit{(3)~Accelerating irregular neural computation in high concurrency.} At Baidu, users raise hundreds of millions of recommendation requests per second. These request executions are irregular because divergent user profiling drives distinct search paths, where the visited items should be estimated for their relevance by the neural metric. Baidu has developed a GPU-accelerated framework for web-scale recommendations to leverage the computational economy of GPUs~\cite{liu2022lion}. However, their performance degrades under highly concurrent, non-uniform workloads because of frequent kernel launches, a process with fixed latency overhead that dominates runtime when computations are small and irregular~\cite{el2016klap}. Millions of sequential processes of deep relevance computations on a GPU cluster introduce bottlenecks of recommendation throughput. Therefore, the third challenge is redesigning computation scheduling to accelerate large-scale irregular neural computations for higher throughput.

To this end, we propose a \textbf{GPU-accelerated Multi-relational Parallel Graph Retrieval~(GMP-GR)} framework for effective and efficient retrieval in web-scale recommendations.
Specifically, we first devise a \textit{Multi-relational User-Item Relevance Metric Learning~(MUIRML)} to generate a practical DNN-based metric via multi-objective optimization.
The metric extracts multi-relational knowledge in inter-dependent user behaviors to capture nuanced user preferences for accurate relevance measurement.
We also infuse our metric's approximate triangle inequality property via a self-covariant loss to navigate large graphs with shorter paths.
Second, we design a \textit{Hierarchically Parallel Approximate Nearest Neighbor Search~(HiPANNS)} algorithm to achieve timely and accurate retrieval for highly concurrent queries.
HiPANNS broadens the search space of the large graph item base by divergent parallel and co-optimized search processes, which can efficiently generate more relevant candidates in real-time retrieval.
Based on it, HiPANNS accelerates irregular neural computations from asynchronous requests by adaptive aggregation in a GPU-based cluster for higher retrieval throughput.
In Baidu recommendation applications, we further apply several system optimization strategies, including traffic dispersal, compilation optimization, model quantization, and timely stream update.
Extensive experiments demonstrate the superior retrieval effectiveness and efficiency of GMP-GR in offline retrieving evaluations and online recommendation services.
Our framework has been deployed in more than twenty web-scale recommender systems at Baidu, where it efficiently handles over one hundred billion requests per day, improving more than $3\%$ time of page for hundreds of millions of users.

The main contributions of our work are concluded as follows:
(1)~We propose a GPU-accelerated Multi-relational Parallel Graph Retrieval framework to achieve accurate and efficient retrieval for web-scale recommendations.
(2)~We devise a DNN-based metric to estimate user-item relevance via capturing inter-dependency among complex user-item relations, which can also improve the search efficiency on a large-scale graph for shorter navigation.
(3)~We design a GPU-accelerated parallel graph-based ANNS algorithm for highly concurrent retrieval by billion-scale breadth-depth balanced graph search and parallel irregular computations with adaptive aggregation.
(4)~We conduct extensive experiments to demonstrate retrieval effectiveness and efficiency improvements in offline and online recommendations at Baidu.

\section{Preliminaries}

Baidu has developed a recommendation framework as a comprehensive funnel mechanism to efficiently retrieve and rank items for users from a web-scale database for over twenty internet products~(e.g., news feed, short video clips, search engine, and online advertising)~\cite{liu2021jizhi}.
The recommendation framework has served over one hundred billion requests for online recommendations made by hundreds of millions of users per day.
Next, we detail the retrieval framework for Baidu recommendations, including the retrieval problem and the retrieval pipeline.

\subsection{Online Retrieval at Baidu}
Given $N_u$ users $\{u_1,u_2,\cdots,u_{N_u}\}$ and $N_v$ items $\{v_1,v_2,\cdots,v_{N_v}\}$. Consider a query from a user $u$ with raw feature $x_{u}$, the objective of the retrieval stage is to recommend the top-$K$ items that are most relevant to $u$ in order.
Specifically, we define the retrieval problem as a function $\mathcal{F}(\cdot;\cdot)$ generates a set $V_{u}$ based on the user features. $V_u$ contains the $K$ most similar items $V_{u} = \{v_{u,1},v_{u,2},\cdots,v_{u,K}\}$ as
\begin{equation}
    V_{u} = \mathcal{F}(x_{u};\{v_1,v_2,\cdots,v_{N_v}\}).
\end{equation}


\subsection{Retrieval Pipeline at Baidu}

Baidu has constructed a graph retrieval framework in their recommendation system, including two stages: offline training and online search~\cite{tan2020fast,chen2022approximate}.

\subsubsection{Step 1. Offline training.}
The first stage aims to learn user and item embeddings associated with a relevance metric for estimating the relevance between users and items.
Embeddings and the metric are learnable modules stored by the system, which are immutable in the search stage.
We can also periodically re-train the modules once the properties change significantly.
Overall, the pipeline of the training step is presented below.

\textit{(1)~Users and item feature construction.}
We first prepare the raw features of the users and items that are denoted as $\{x_{u_1},x_{u_2},\cdots,x_{N_u}\}$ and $\{x_{v_1},x_{v_2},\cdots,x_{N_v}\}$, where $N_u$ and $N_v$ are number of users and items, respectively. Meanwhile, the interaction records of users $u$ and items $b$ are denoted as $\{D^z_{u,v}\}$. For an interaction type $z$, $D^z_{u_i,v_j}$ is the interactive value between a user $u_i$ and an item $v_j$, which reflects how close they are in terms of the interaction $z$.

\textit{(2)~User and item embedding projection.}
Then, we pre-define a user and an item projector, $\mathcal{H}_u(\cdot)$ and $\mathcal{H}_v(\cdot)$, to transform raw features into learnable embeddings as user and item representations, which are defined as $h_{u_i} = \mathcal{H}_u(x_{u_i})$ and $h_{v_j} = \mathcal{H}_v(x_{v_j})$.

\textit{(3)~User-item relevance metric definition.}
After that, we devise a DNN-based learnable metric~\cite{dehghani2017neural,tay2018latent} to estimate the relevance between a user and an item by a high-dimensional calculation of their embeddings.
Hence, we aim to obtain a metric $\Delta(\cdot,\cdot)$ to quantify the relevance value given a user embedding and an item embedding as
\begin{equation}
    \delta_{u,v} = \Delta(h_{u},h_{v}).
\end{equation}

\textit{(4)~Optimization of embeddings and the metric.}
Next, we optimize the user embeddings, item embeddings, and the relevance metric based on the user features, item features, and user-item interactions. We describe the optimization as a function $\mathcal{F}_t(\cdot,\cdot,\cdot)$ that
\begin{equation}
    \{h_u\}, \{h_v\}, \Delta(\cdot,\cdot) = \mathcal{F}_t(\{x_u\}, \{x_v\}, \{D^z_{u,v}\}).
\end{equation}
The learned embeddings and the relevance metric are used in the online search stage to measure user-item relevance. 

\begin{figure*}
\setlength{\abovecaptionskip}{0.1cm}
\setlength{\belowcaptionskip}{-0.3cm}
    \centering
    \includegraphics[width=\linewidth]{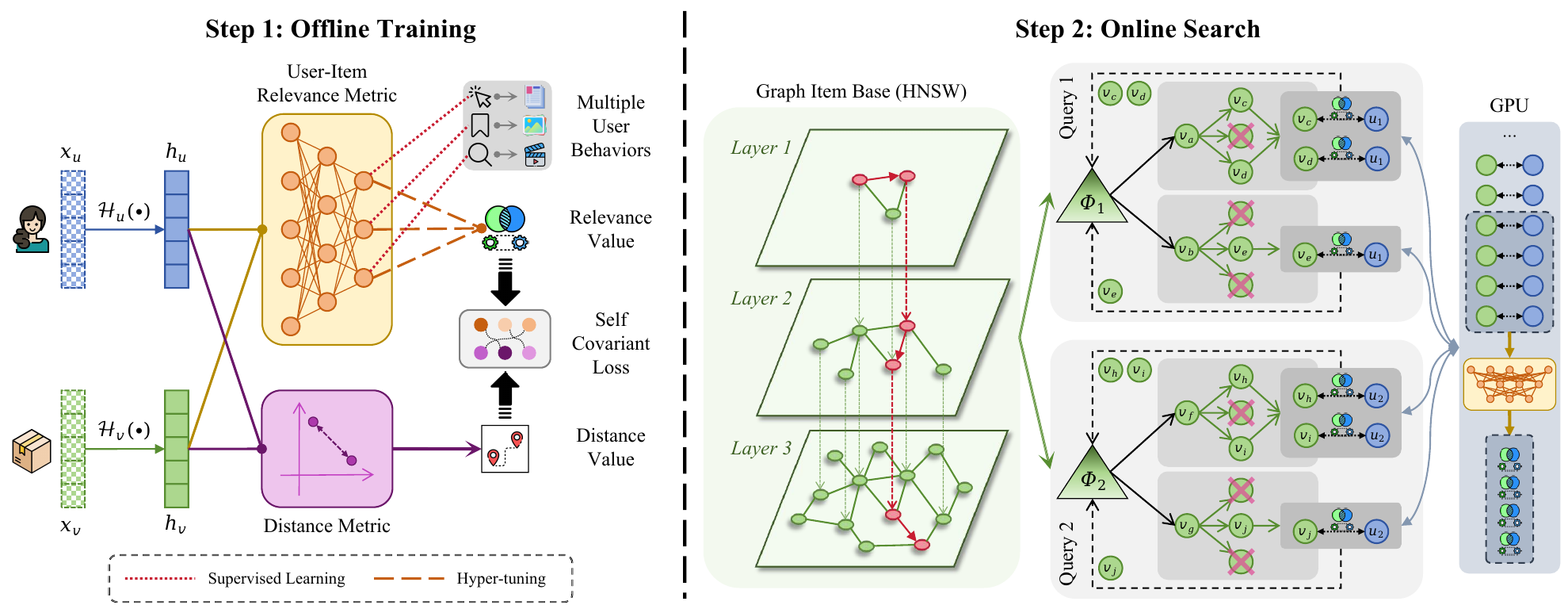}
    \caption{An overview of the GPU-accelerated Multi-relational Parallel Graph Retrieval framework.}
    \label{fig:gmp-gr}
\end{figure*}

\subsubsection{Step 2. Online search.}
In the second stage, we construct a graph-based item database that supports an efficient search for the target items.
Traditional tree-based ANNS fails to handle web-scale data with high dimensions and uneven distributions, and requires extensive tree construction costs.
Instead, we follow graph-based ANNS to construct a graph item base that can efficiently navigate possibly relevant items and retrieve satisfactory items for given users based on graph search.
Specifically, we adopt HNSW~\cite{malkov2018efficient} as the graph structure.
HNSW utilizes a multi-layered graph structure, denoted as a $L$-layer graph $G$, where each upper layer contains fewer nodes to facilitate efficient search paths. Based on its hierarchical nature, we can effectively narrow the search space for nearest neighbors in the base layer, initiating the search process from the upper layers.

In the online search, we infer $K$ relevant items with a querying user as follows. We perform the search iteratively across each layer of the HNSW structure, with the selected items from one layer serving as seed nodes for traversal in the subsequent layer. For each layer, we first access the corresponding user embedding. Second, we generate seed nodes, from which we start to traverse the graph. Third, we expand the candidate node set by visiting neighbors of existing nodes and continuously update the most similar ones for the queried user. The $K$ most similar items are ultimately extracted from the final layer.
The overall search process for the relevant item set $\hat{V}_u$ for user $u$ can be defined as
\begin{equation}
    \hat{V}_u = \mathop{\arg\max}\limits_{V_u \in V} \sum\limits_{v \in V_u} \delta_{u,v}.
\end{equation}

\section{GPU-accelerated Multi-relational Parallel Graph Retrieval Framework}

This section introduces the GPU-accelerated Multi-relational Parallel Graph Retrieval~(GMP-GR) framework deployed at Baidu for web-scale recommendation.
As shown in Figure~\ref{fig:gmp-gr}, GMP-GR equips two main modules to achieve effective and efficient performance.
In the training step, a multi-relational user-item relevance metric learning generates a DNN-based metric capturing inter-dependent relationships between users and items and inherits the triangle inequality property for path-efficient graph search.
In the search step, a hierarchically parallel graph-based ANNS algorithm effectively delivers reliable search candidates in high concurrency, which explores a breadth-depth balanced search space and synchronizes large-scale irregular neural computations on GPUs.

\subsection{Multi-relational User-item Relevance Metric Learning}\label{sec:metric}

The training step aims to optimize the user and item embeddings with the DNN-based metric to model the relevance between users and items.
However, existing DNN-based metrics either fail in insufficient understanding of nuance user preference due to partial incorporation of multi-dimensional behaviors on items, or do not satisfy the triangle inequality property necessitated by efficient navigation of graph-based ANNS.
Therefore, we propose a \textit{Multi-relational User-Item Relevance Metric Learning~(MUIRML)} algorithm to train a relevance metric that not only ensures elaborate multi-relational modeling, but also enhances navigating efficiency in the search step, ultimately improving the retrieval effectiveness.

\textit{(1)~DNN-based Relevance computation.}
We first utilize a Multi-Layer Perceptron~(MLP) to model multiple interactive behaviors.
Initially, the input consists of the user embedding $h_u$, the item embedding $h_v$, and the serendipity vector $\tilde{h} \sim \mathcal{N}(0,1)$ for diverse exploration as
\begin{equation}
    h_{u,v}^{(0)} = \operatorname{concat}(h_u+\tilde{h},h_v),
\end{equation}
the MLP inputs $h_{u,v}^{(0)}$ and aims to output the interaction vector $Z_{u,v}$ that stacks multiple interactions $z_{u,v}$ between user $u$ and item $v$.
The overall operation can be defined as
\begin{equation}
    \hat{Z}_{u,v} = \operatorname{MLP}(h_{u,v}^{(0)}),
\end{equation}
where $\hat{Z}_{u,v}$ is the predicted interaction vector.
Specifically, we define the computation of the $m$-th layer MLP as
\begin{equation}
    h_{u,v}^{(m)} = \mathbf{W}^{(m)} \cdot h_{u,v}^{(m-1)}+b^{(m)},
\end{equation}
where $h_{u,v}^{(m)}$ is the $m$-th layer joint representation of user $u$ and item $v$, and $\mathbf{W}^{(m)}$ and $b^{(m)}$ are the $m$-th layer learnable parameters.

Then, we leverage an attention mechanism to weighted aggregate predicted interaction values for a joint understanding of diverse user behaviors.
Specifically, the attention vector $\mathbf{A}$ is multiplied on $\hat{Z}_{u,v}$ to produce the final user-item relevance value $\delta_{u,v}$ as
\begin{equation}
    \delta_{u,v} = \mathbf{A} \cdot \hat{Z}_{u,v},
\end{equation}
where $\mathbf{A}$ is initialized with $\frac{1}{|\mathbf{A}|}$ and can be hyper-tuned to achieve personalized behavior importance for different web recommendation scenarios.

\textit{(2)~Self-supervised loss for short graph navigation.}
Instead of directly forcing the alignment between user-item relevance values computed by the DNN-based metric and the Euclidean distance, we aim to retain the triangle inequality property by transplanting the ranking regularity of Euclidean distance into the DNN-based metric.
Specifically, we devise a Self Covariant Loss~(SCL) to quantify the gap of the ranking regularity via the divergence of the pairwise values covariance for two metrics.
By minimizing the loss, the order of items ranked according to our metric will be similar to the Euclidean distance, while the relevance values will be maximally retained for accurate estimation by multi-dimensional understanding.
In this way, we can eliminate unnecessary traversal paths and increase the possibility of approaching the optimal candidates.

Specifically, we randomly sample $N_l$ pairs of users and items $\{(u_{i_1},v_{j_1}),\cdots,(u_{i_{N_l}},v_{j_{N_l}})\}$ and compute their DNN-based relevance values $\{\delta_{u,v}\}$ and Euclidean distance $\{\delta^{\prime}_{u,v}\}$, respectively. Then, we calculate the correlation of two groups of values based on their covariance as
\begin{equation}
    \rho_{u,v} = {\frac {\sum \limits _{k=1}^{N_l}(\delta_{u_k,v_k}-{\overline {\delta_{u,v}}})(\delta^{\prime}_{u_k,v_k}-{\overline {\delta^{\prime}_{u,v}}})}{{\operatorname{VAR}(\delta_{u,v})} \cdot {\operatorname{VAR}(\delta^{\prime}_{u,v})}}},
\end{equation}
where $\operatorname{VAR}(x)=\sqrt {\sum_{k=1}^{N_l}(x-{\overline {x}})^{2}}$.

\textit{(3)~Multi-objective optimization for embeddings and the relevance metric.}
Lastly, we design the multi-objective optimization to train the user and item embeddings, as well as the DNN-based relevance metric. The loss is defined as
\begin{equation}
    \mathcal{L} = \sum\limits_{i=1}^{N_u} \sum\limits_{j=1}^{N_v} (\Vert Z_{u_i,v_j} - \hat{Z}_{u_i,v_j} \Vert_2 + \sum\limits_{k=1}^{N_l} \frac{1}{\rho_{u_{i_k},v_{j_k}}+1}),
\end{equation}
which will be optimized by gradient-based algorithms.
In this way, we can simultaneously minimize the prediction error of multi-relational interaction behaviors and maximize the similarity of ranking regularity between our DNN-based metric and Euclidean distance for triangle inequality property, which can generate shorter paths in graph search for more efficient retrieval.

\subsection{GPU-accelerated Hierarchically Parallel Approximate Nearest Neighbor Search}\label{sec:search}

In this work, the search step aims to retrieve the most similar items using graph-based ANNS.
Existing graph retrieval frameworks struggle to achieve low-latency retrieval in high-concurrency scenarios for two reasons.
First, their traversal strategies underperform at scale, as they inadequately balance search depth, \ie refinement of local neighborhoods, and breadth, \ie coverage of global graph regions, within constrained computational budgets~\cite{ootomo2024cagra}.
Second, the underlying GPU-based infrastructure faces inefficiencies when handling asynchronous retrieval requests, which includes tremendous irregular neural computations according to divergent search routes. The frequent kernel launch of GPU computation from different queries induces significant overhead and contention under extreme concurrency~\cite{pan2023recom}.
These limitations collectively degrade the retrieval throughput for real-time recommendations.
To this end, we propose a \textit{Hierarchically Parallel Approximate Nearest Neighbor Search~(HiPANNS)} algorithm to efficiently answer highly concurrent retrieval queries.
Specifically, HiPANNS consists of two sub-algorithms, \ie inter-candidate HiPANNS and inter-query HiPANNS.

Inter-candidate HiPANNS~(C-HiPANNS) allows parallel search within the large-scale graph item base to broaden the global search space for more comprehensive retrieval.
For each query $u$, we aim to execute $k$ parallel search processes with different initial seeds $S$.
Each search process traverses the graph item base by visiting the most relevant neighbors.
We jointly maintain candidates by a shared max heap $\Phi_u$ that ranks $k$ current most relevant items among processes.
In this work, we carry out the multi-processing at each layer of the HNSW graph, seeded with top-$k$ candidates ranked by $\Phi_u$ of the last layer.
The parallelized traversal achieves a $k$-fold expansion of the search space, which yields more effective retrieval results.

Inter-query HiPANNS~(Q-HiPANNS) adaptively aggregates irregular neural computations for more efficient GPU-accelerated retrieval against launch bound from extreme concurrency.
We adaptively synchronize the irregular user-item relevance evaluation by a queuing aggregation scheme and parallelize batched neural computation on GPU for acceleration.
First, we initialize a queue $\Psi$ to persistently listen to neural computations. $\Psi$ will record the computations orderly for the following sequential batching, \ie the element received first will be batched first. Each element in $\Psi$ is produced from C-HiPANNS as a group of $(u,v)$ pairs for measuring neighbor embedding similarity. We dynamically generate batches for parallel computing the GPU cluster $U$.
However, the elements of $\Psi$ vary in amounts due to divergent connections in the graph item base, causing difficulties in aggregating fixed batches that are beneficial to fully utilize the computational ability of GPUs.
To address that, we synthesize batches by adaptively slicing $(u,v)$ groups to fit certain configurations.
Specifically, we set the maximum size of the computing batch as $B_U$. If the size of current batch $B_i$, $N_{B_i}$, is larger than the remaining space $\epsilon$, we split it into two elements, including $B_i^{[1,\epsilon]}=\{(u_{i,1},v_{i,1}),\cdots,(u_{i,\epsilon},v_{i,\epsilon})\}$ and $B_i^{[\epsilon+1,N_{B_i}]}=\{(u_{i,\epsilon+1},v_{i,\epsilon+1}),\cdots,(u_{i,N_{B_i}},v_{i,N_{B_i}})\}$. $B_i^{[1,\epsilon]}$ will be computed in the current executive round of $U$ while $B_i^{[\epsilon+1,N_{B_i}]}$ will be computed in the next.
The output values will be returned as estimations of the relevance between users and items during the processes of C-HiPANNS.

More details are presented in Appendix~\ref{sec:appendix_pseudocode}.

\section{System Optimization}\label{sec:sys_op}

We devise various system optimization strategies to accelerate our retrieval framework deployed at Baidu.

\subsection{Traffic Dispersal}
We construct a communicative computing cluster composed of multiple interconnected servers designed to facilitate traffic dispersal and mitigate the effects of extreme query loads.
Each server is equipped with a monitoring module that continuously assesses its computing load by analyzing available memory and current query traffic levels.
When a server reaches its saturation point, it proactively requests assistance from its peer servers to redistribute the excess queries. The receiving servers will accept these queries if they possess adequate computational resources at that time.
This approach significantly enhances overall system efficiency by optimizing the utilization of distributed computational resources.

\subsection{Compilation Optimization}
Metric learning and graph search algorithms employ fundamental operators.
To accelerate retrieval processes, we integrate these operators to minimize redundant computations based on existing techniques of computational graph optimization and pattern recognition~\cite{niu2021dnnfusion}.
Our system includes an automatic analysis module for operator kernel compilation, which aids in the optimization of neural computations for specific hardware configurations.
Moreover, to address the launch bound issue that may affect computational performance on GPUs~\cite{wang2014characterization}, we abstract the execution process of DNN-based operations as a graph. This approach utilizes an adaptive bucketing mechanism to aggregate independent submissions of individual operations into a single, cohesive submission, thereby enhancing computational efficiency.

\subsection{Model Quantization}
We utilize a post-training quantization technique on the DNN-based relevance metric to decrease the computing costs with little accuracy loss. Specifically, we reduce the size of the entire MLP connection layers by quantizing the weights from $32$-bit floating point numbers to $16$-bit. After reductions, the metric only requires half of the size for computing and memorization and supports faster operators tailored for lower precise calculation. In large-scale recommendations, this method can reduce retrieval latency and significantly conserve computational resources on participating processing units for higher throughput.

\subsection{Timely Stream Update}
The user and item information is time-varying. To deliver accurate recommendations over time, it is crucial for the system to maintain up-to-date user and item representations and model parameters to reflect the evolving knowledge.
However, increasing the freshness of this information necessitates more frequent updates, which can lead to higher training costs.
To balance freshness and training costs, we first design a stepwise entry scheme to select active users and items via a dynamic threshold for timely updates.
Besides, we use a click trigger that activates updates of the item base to incorporate items that occurred initially.
Fresh data is then cumulatively integrated with the original item base in a batched manner to ensure both responsiveness and efficiency in handling data updates.

\begin{table*}[t]
\centering
\caption{Offline retrieval performance of different methods on four datasets for three metrics.}
\label{tab:overall}
    \begin{tabular}{@{}c|ccc|ccc|ccc|ccc@{}}
    \toprule
           & \multicolumn{3}{c|}{Video} & \multicolumn{3}{c|}{MVideo} & \multicolumn{3}{c|}{Explore} & \multicolumn{3}{c}{Fresh} \\
           \midrule
           & Cov    & Rec@10 & Rec@100 & Cov     & Rec@10 & Rec@100 & Cov     & Rec@10  & Rec@100 & Cov    & Rec@10 & Rec@100 \\
    \midrule
    C-NSW & 0.3386 & 0.3829 & 0.4028  & 0.3185  & 0.3775 & 0.3986  & 0.3288  & 0.3486  & 0.3674  & 0.3493 & 0.3823 & 0.4084  \\
    D-NSW    & 0.4084 & 0.5527 & 0.6094  & 0.4283  & 0.5428 & 0.5908  & 0.4081  & 0.5265  & 0.5376  & 0.5083 & 0.6075 & 0.6344  \\
    SL2G   & 0.4413 & 0.5784 & 0.6121  & 0.4428  & 0.5684 & 0.6049  & 0.4076  & 0.5249  & 0.5485  & 0.5123 & 0.6285 & 0.6298  \\
    NANN   & 0.5368 & 0.6717 & 0.6635  & 0.5021  & 0.6491 & 0.6498  & 0.5029  & 0.6038  & 0.6345  & 0.6130 & 0.7318 & 0.7555  \\
    \midrule
    GMP-GR  & \textbf{0.6492} & \textbf{0.8239} & \textbf{0.8052}  & \textbf{0.6044}  & \textbf{0.7175} & \textbf{0.7325}  & \textbf{0.6279}  & \textbf{0.7999}  & \textbf{0.7810}  & \textbf{0.6906} & \textbf{0.8420} & \textbf{0.8428} \\
    \bottomrule
    \end{tabular}
\end{table*}

\section{Experiments}\label{sec:exp}

This section introduces the experiments of GMP-GR under offline and online settings. Our goal is to answer the following questions through the investigation of results.
Specifically, for offline experiments,
(1)~How accurate is GMP-GR in retrieving precisely relevant items?
(2)~How does the efficiency of GMP-GR compare to baseline retrieval frameworks?
(3)~What is the effectiveness of MUIRML and HiPANNS within GMP-GR?
For online experiments, how does GMP-GR affect real-world recommendation statistics?

\subsection{Experimental Setup}

We present datasets, baselines, and metrics in our offline and online experiments. Besides, implementation details are provided in Appendix~\ref{sec:appendix_implementation}.

\subsubsection{Datasets.}\label{sec:dataset}
We collect real-world video resources from Baidu applications and construct four datasets according to their potential of user satisfaction, including (1)~Video: popular landscape videos, (2)~MVideo: popular portrait videos, (3)~Explore: videos that are potentially favorable but not popular, and (4)~Fresh: videos that are uncommon and infrequently discovered.

\subsubsection{Baselines.}
We adopt four retrieval baselines in offline experiments, including
(1)~C-NSW: a basic retrieval framework with cosine function as the relevance metric based on NSW~\cite{malkov2014approximate};
(2)~D-NSW: a DNN-based retrieval framework based on NSW, supervised by single user-item interaction;
(3)~SL2G~\cite{tan2020fast}: a DNN-based retrieval framework based on HNSW~\cite{malkov2018efficient};
and (4)~NANN~\cite{chen2022approximate}: an industrial retrieval framework improved on SL2G with adversarial training and more efficient ANNS.

\subsubsection{Metrics.}
For offline experiments, we involve three metrics in our evaluations on retrieval performance, including Coverage~(Cov), Recall@10~(Rec@10), and Recall@100~(Rec@100).
We take the brute-force-sorted items as the ground truths of retrieval.
Cov describes the percentage of retrieved items that rank in the top $100$ items of the ground truths.
Rec@10 and Rec@100 measure the percentage of the top $10$ and $100$ items of the ground truths in the top $10$ and $100$ retrieved items, respectively.
Moreover, we evaluate the speed by the number of items retrieved per second. For online testing, we collect query traffic, latency, and throughput of servers for different services across timestamps.

\subsection{Offline Evaluation}

We evaluate the retrieval performance and efficiency of GMP-GR and baseline frameworks on four datasets in an offline environment.

\subsubsection{Retrieval performance.}
Overall, GMP-GR consistently outperformed the baseline methods across all settings. For instance, GMP-GR demonstrated improvements of at least $20.94\%$, $22.66\%$, and $21.36\%$ in terms of Cov, Rec@10, and Rec@100, respectively, on the Video dataset. Notably, significant improvements are also evident on other datasets.
Specifically, compared to SL2G and NANN, two typical industrial retrieval frameworks consistently based on HNSW, GMP-GR produces more accurate retrieved results. GMP-GR achieved performance improvements of $20.37\%$, $24.86\%$, and $12.66\%$ over SL2G and NANN on Cov on MVideo, Explore, and Fresh, respectively. The potential reasons are that~(1)~GMP-GR overcomes the shortage of SL2G that the neural metric may harm graph search efficiency due to the unsatisfaction of triangle inequality, and (2)~GMP-GR incorporates valuable insights from dimensional interactive information in metric learning and leverages a broader search space for higher possibility of effective retrieval.

\begin{figure}[t]
    \centering
    \subfigure[Video]{\includegraphics[width=0.48\linewidth]{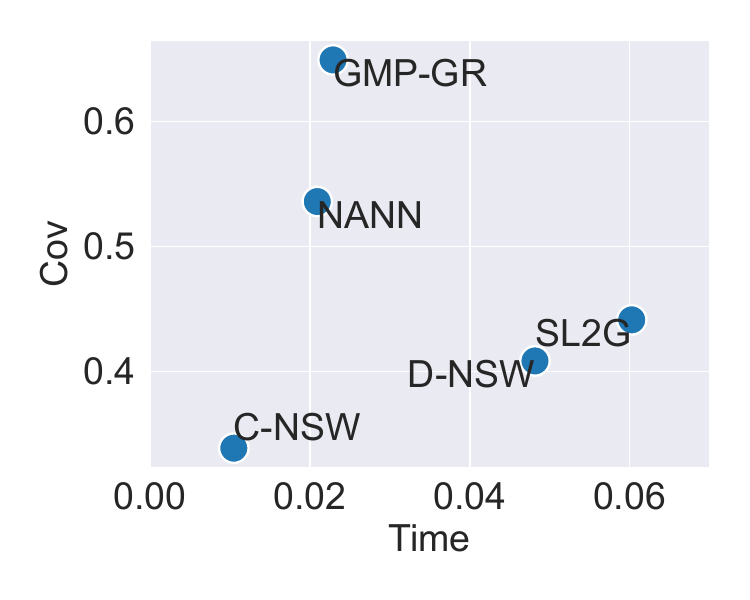}\label{fig:effciency_video}}
    \subfigure[Explore]{\includegraphics[width=0.48\linewidth]{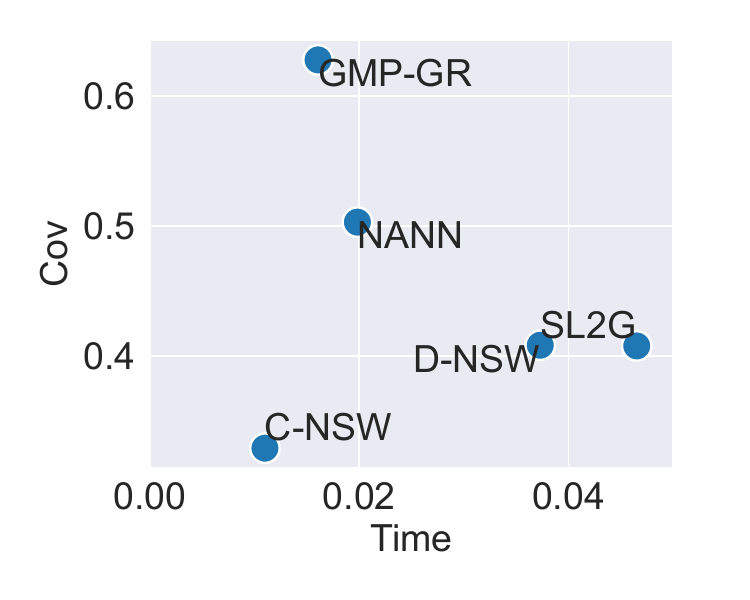}\label{fig:effciency_explore}}
    \caption{Analysis of correlations between retrieval performance and efficiency on Video and Explore datasets. Methods close to the left~top~($\nwarrow$) have better retrieval performance, \ie Cov~($\uparrow$), with less time costs~($\leftarrow$).}
    \label{fig:efficiency}
\end{figure}

\begin{table}[t]
    \centering
    \caption{Retrieval speed (items per second).}
    \label{tab:speed}
    \begin{tabular}{@{}c|cccc@{}}
        \toprule
        & Video & MVideo & Explore & Fresh\\
        \midrule
        C-NSW & 520,390 & 405,096 & 521,639 & 345,577 \\
        D-NSW & 163,093 & 163,821 & 199,163 & 170,889 \\
        NANN & 281,758 & 226,280 & 248,118 & 150,022 \\
        SL2G & 134,947 & 145,312 & 162,033 & 176,316 \\
        GMP-GR & 321,352 & 299,242 & 384,458 & 229,902 \\
        \bottomrule
    \end{tabular}
\end{table}

\subsubsection{Efficiency analysis.}
This part analyzes the efficiency by testing their inference time and search speed.
We first analyze the correlation between retrieval performance and time by scattering five methods in Figure~\ref{fig:efficiency}. More results on MVideo and Fresh datasets are in Appendix~\ref{sec:appendix_efficiency}.
Intuitively, a method is considered more efficient if it achieves higher performance~(here we use Cov to evaluate) with lower time consumption. Results reveal that GMP-GR outperforms the four baseline methods in terms of efficiency, as evidenced by its proximity to the upper left quadrant in the scatter plots.
Notably, C-NSW stands out for its minimal retrieval time, attributed to its reliance on the cosine similarity metric which is computationally lightweight than DNNs. However, C-NSW underperforms in retrieval accuracy due to the limitation that the simple metric is unable to capture complex user-item relationships.
Then, we compare the retrieval speed in Table~\ref{tab:speed}, which reflects the number of nodes that methods traverse on HNSW per second. GMP-GR also exhibits the capability to search through more nodes~(\ie larger search space) than any DNN-based approach.
In summary, GMP-GR successfully defeats all baseline methods, striking an optimal balance between retrieval efficiency and accuracy.

\begin{figure}[t]
    \centering
    \subfigure[Cov]{\includegraphics[width=0.48\linewidth]{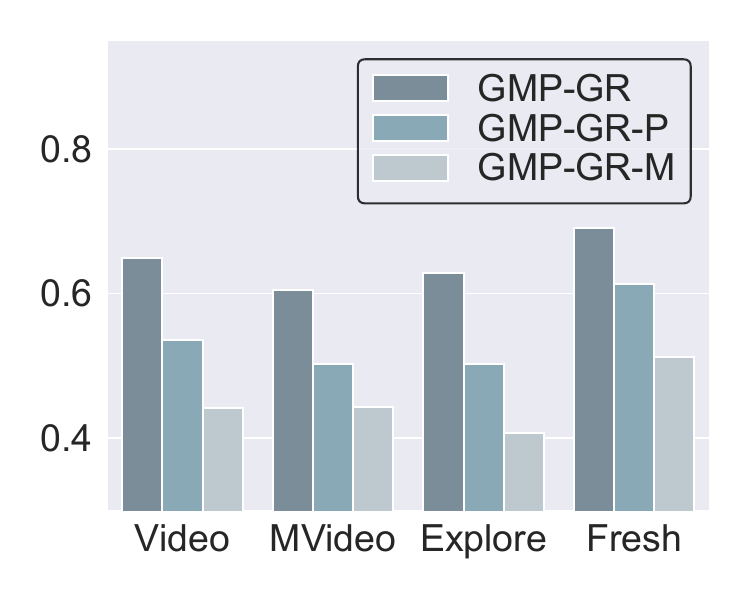}\label{fig:ablation_cov}}
    \subfigure[Speed~($10^6$ items / sec)]{\includegraphics[width=0.48\linewidth]{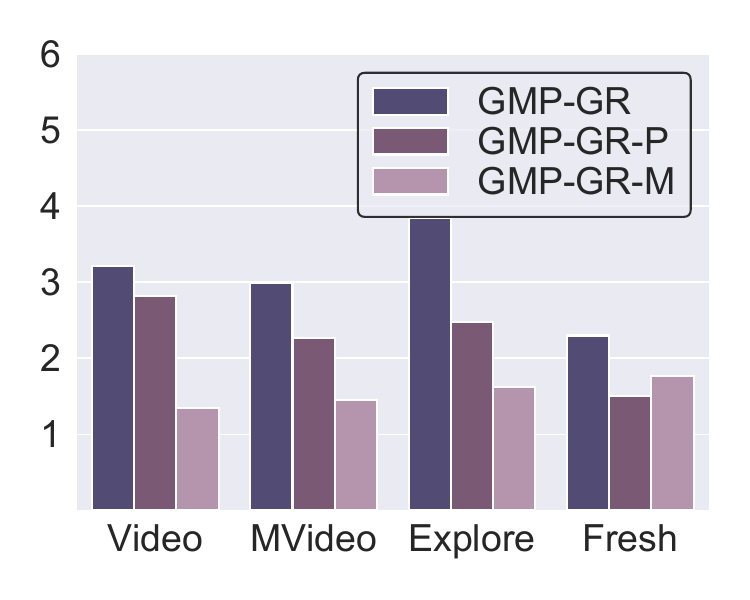}\label{fig:ablation_item_per_s}}
    \caption{Ablation study of GMP-GR on retrieval performance~(Cov) and speed.}
    \label{fig:ablation}
\end{figure}

\subsubsection{Ablation study.}
We conduct an ablation study on GMP-GR in Figure~\ref{fig:ablation}, by comparing it with two variants, \ie GMP-GR-P and GMP-GR-M. Evaluations on Rec@10 and Rec@100 are provided in Appendix~\ref{sec:appendix_ablation}.
GMP-GR-P replaces MUIRML with a single supervision of the behavior in which a user visits an item, and GMP-GR-M substitutes HiPANNS with a standard HNSW search algorithm.
Specifically, we find that GMP-GR consistently outperforms both GMP-GR-P and GMP-GR-M across different datasets.
This superiority can be attributed to two key factors. First, MUIRML significantly enhances the relevance estimation accuracy and pathfinding efficiency. Second, HiPANNS prevents the graph-based ANNS process from succumbing to local optima by distributed and co-optimized search.
In addition, GMP-GR also shows a notable advantage in retrieval speed that replacing HiPANNS with a standard HNSW search algorithm can result in a great degradation of retrieval speed. This observation underscores the pivotal role of HiPANNS in acceleration. Furthermore, a more effective user-item relevance metric not only enhances the accuracy but also reduces traversal steps to converge to optimal candidates.

\begin{table}[t]
    \caption{Online user statistics for two services.}
    \label{tab:user_exp}
    \centering
    \begin{tabular}{c|cc|cc}
        \toprule
        & \multicolumn{2}{c|}{Time on Page} & \multicolumn{2}{c}{Daily Active Users} \\
        & \textit{Service 1} & \textit{Service 2} & \textit{Service 1} & \textit{Service 2} \\
        \midrule
        GMP-GR & +1.13\% & 1.19\% & +0.23\% & +0.50\% \\
        \bottomrule
    \end{tabular}
\end{table}

\subsection{Online Testing}

We deploy GMP-GR in the Baidu Mobile APP for two services supported by five servers for online testing.
Service 1 is for the main recommendation list on the initial page.
Service 2 is for the next recommendation sequence that can be accessed after clicking on a video in immersive mode.

\begin{figure}[t]
    \centering
    \subfigure[Traffic of two weeks in June, 2024.]{\includegraphics[width=\linewidth]{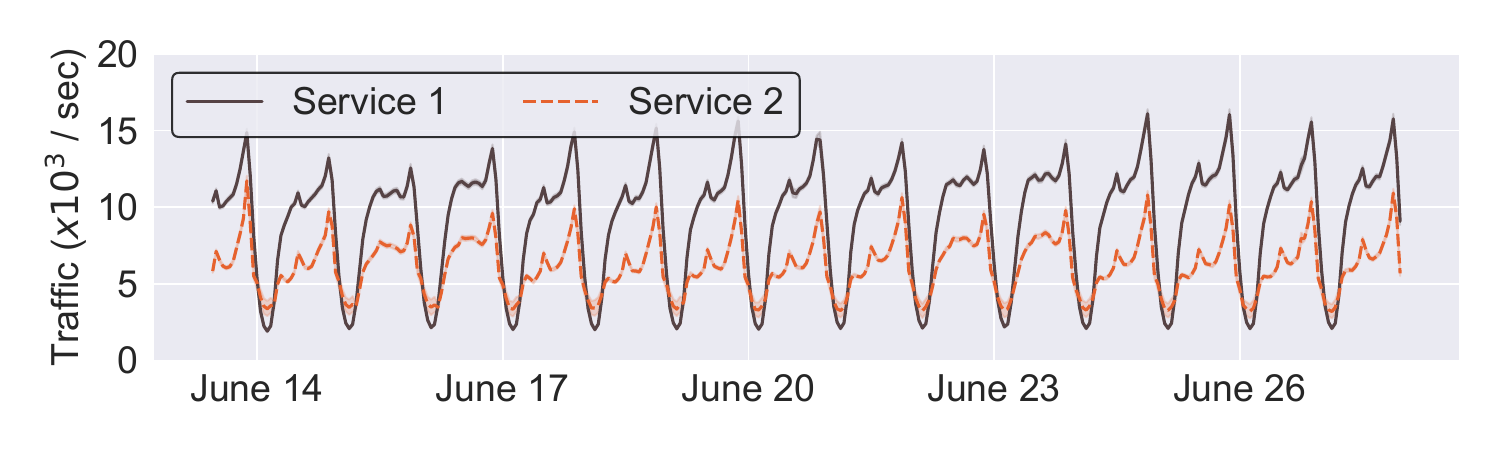}\label{fig:timestamp_traffic}}
    \subfigure[Latency in one day.]{\includegraphics[width=\linewidth]{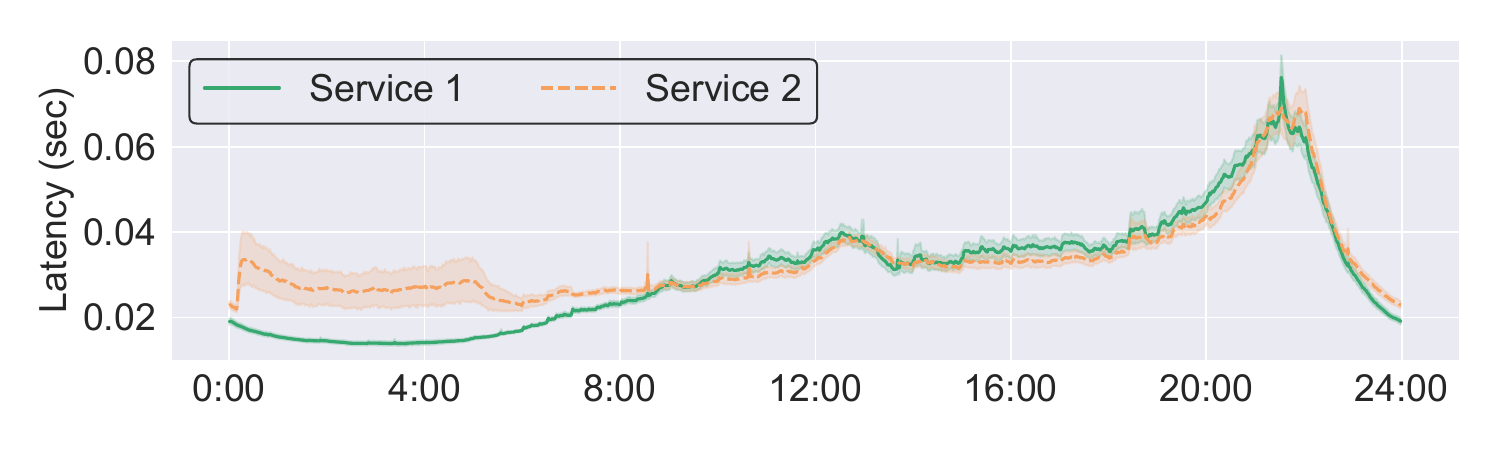}\label{fig:timestamp_latency}}
    \caption{Temporal statistics of traffic and latency.}
    \label{fig:timestamp_traffic_latency}
\end{figure}

\subsubsection{Time on page and daily active users.}
Table~\ref{tab:user_exp} depicts the impact of deploying GMP-GR on time on page and daily active users. Our results indicate a significant improvement in time on page and daily active users across the two services.
The integral components of GMP-GR, MUIRML, and HiPANNS, contribute to a $0.57\%$ and $0.23\%$ improvement in time on page, respectively.
Our findings suggest that the deployment of GMP-GR with improved modules has positively impacted user engagement and satisfaction.

\subsubsection{Query throughput.}
We also examine the query throughput influenced by the system acceleration strategies. The results demonstrate a strong improvement in query throughput, with an approximate doubling of throughput in the two services. Specifically, our analysis shows that the compilation optimization has led to a $74\%$ and $75\%$ increase in throughput by simplifying the computational graphs. Similarly, the model quantization has resulted in a $26\%$ and $25\%$ increase in throughput by decreasing DNN-based calculations. These findings underscore the effectiveness of GMP-GR in optimizing query processing and system efficiency.

\begin{figure}[t]
    \centering
    \subfigure[Service 1]{\includegraphics[width=0.48\linewidth]{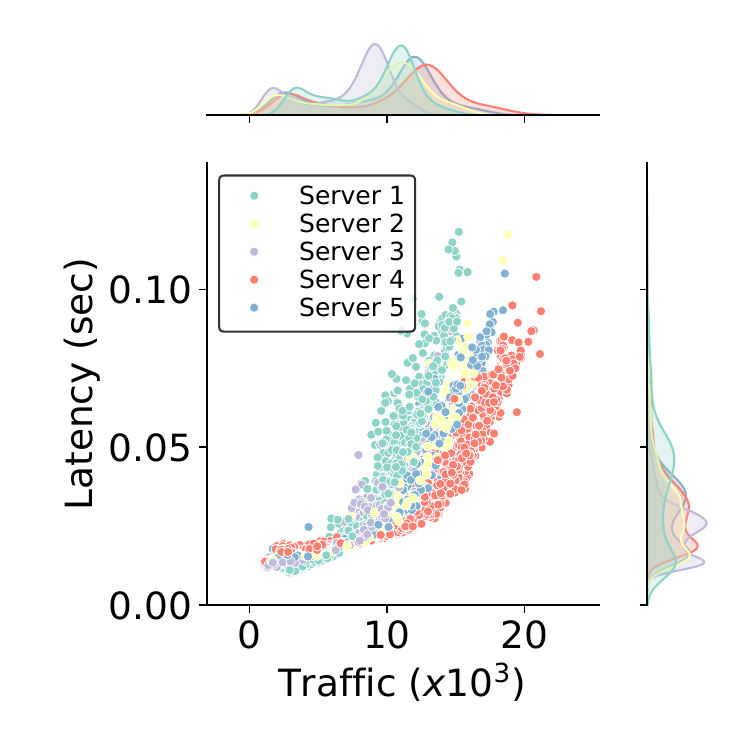}\label{fig:service_1_traffic_latency}}
    \subfigure[Service 2]{\includegraphics[width=0.48\linewidth]{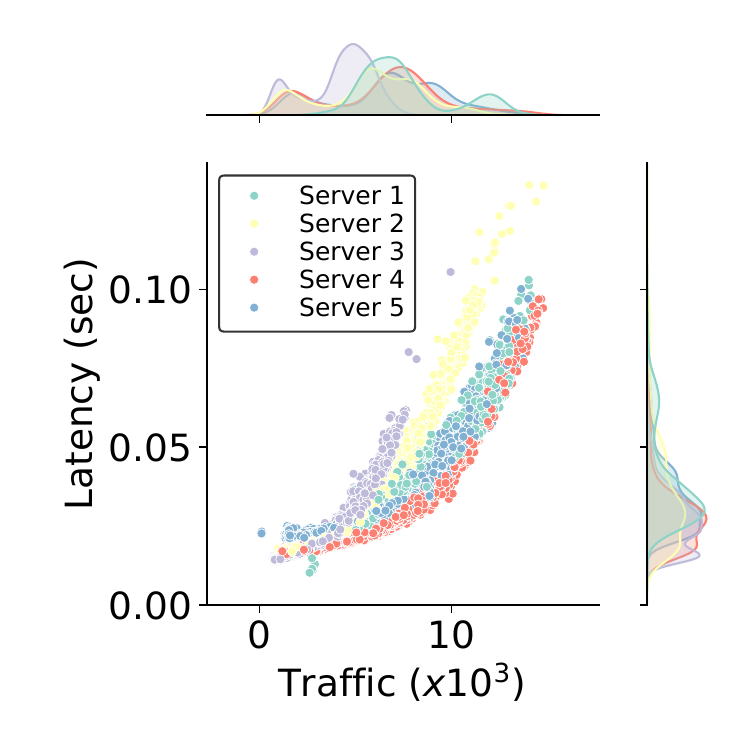}\label{fig:service_2_traffic_latency}}
    \caption{Analysis of distributions of traffic and latency.}
    \label{fig:service_traffic_latency}
\end{figure}

\begin{figure}[t]
    \centering
    \subfigure[Traffic]{\includegraphics[width=0.48\linewidth]{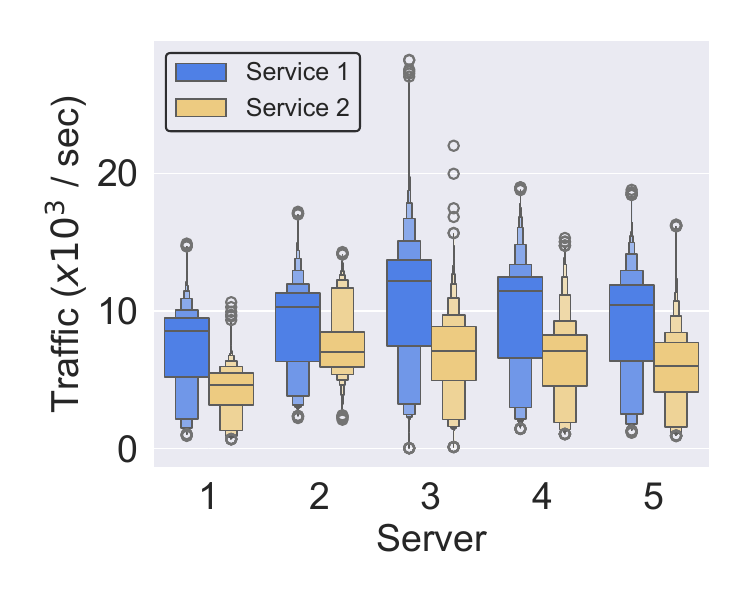}\label{fig:server_traffic}}
    \subfigure[Latency]{\includegraphics[width=0.48\linewidth]{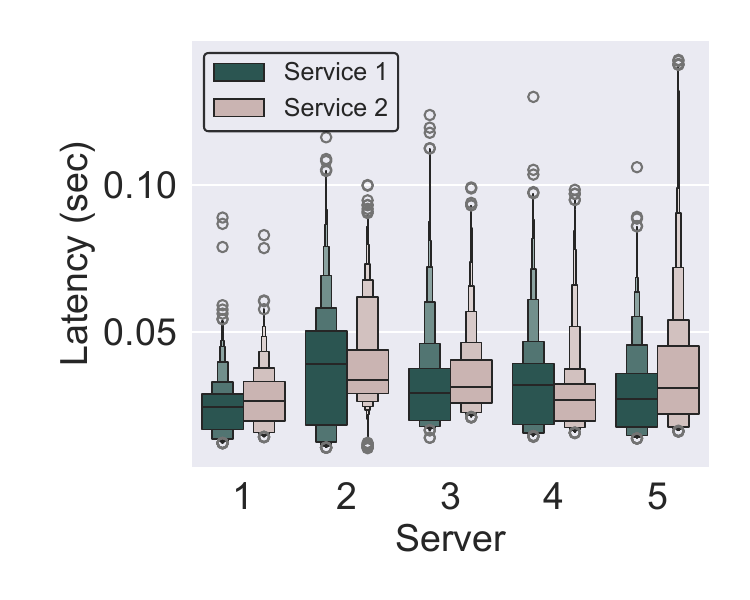}\label{fig:server_latency}}
    \caption{Traffic and latency distributions.}
    \label{fig:server_traffic_latency}
\end{figure}

\subsubsection{Statistical analysis of traffic and latency.}
We collect traffic and latency statistics per second, which are distributedly recorded at five servers in two services.
First, we examine the temporal patterns as shown in Figure~\ref{fig:timestamp_traffic_latency}. Specifically, Figure~\ref{fig:timestamp_traffic} illustrates the traffic trends over a two-week period, highlighting periodic fluctuations in query volumes. Meanwhile, Figure~\ref{fig:timestamp_latency} provides a more granular view of latency statistics over a 24-hour cycle, revealing variations in online response times in one day.
Moreover, we investigate the distributions of traffic and latency and explore their correlations in Figure~\ref{fig:service_traffic_latency}. The result indicates a direct relationship between query traffic and latency, with higher traffic volumes leading to longer response times. Furthermore, we observe distinct patterns in the latency-traffic relationships across five servers, reflecting divergence in computational resources and workload distribution.
Finally, we summarize the traffic and latency statistics across five servers in Figure~\ref{fig:server_traffic_latency}, providing insights into the performance characteristics in terms of query processing and response times.

\section{Related Works}\label{sec:related_work}

In this section, we review the related works for our study, including deep neural network-based recommender systems and approximate nearest neighbor search.

\subsection{Deep Neural Network-based Recommender Systems}
Deep Neural Network~(DNN) has become a popular tool for recommendation systems due to its ability to capture intricate user-item interaction patterns through nonlinear transformations~\cite{zhang2019deep}.
For example, \cite{zhang2017hashtag} leverages convolutional and recurrent neural networks for recommendations to extract features from images and deep sequential models to learn text features from tweets.
Recently, neural collaborative filtering has been increasingly adopted, where a Multi-layer Perceptron~(MLP) is deployed to approximate the interaction function and has significant performance gains over traditional methods~\cite{he2017neural}.
Along with the idea, two-tower architectures have become popular in DNN-based recommender systems~\cite{yang2020mixed}. They train dual encoders for users and items, respectively, which generate user and item representations for downstream operations such as relevance evaluation.
With two sets of high-dimensional embeddings, the DNN-based system can provide recommendations in user-based or item-based paradigms~\cite{ko2022survey}.
User-based collaborative filtering compares similarities between users based on their past ratings and recommends items liked by similar users~\cite{zhang2020alleviating}.
In contrast, item-based collaborative filtering computes similarities between items rather than between users~\cite{xue2019deep}.
We design a DNN-based metric optimized in a multi-objective way to incorporate comprehensive behaviors and triangle inequality for better recommendation accuracy and efficiency.

\subsection{Approximate Nearest Neighbor Search}
Approximate Nearest Neighbor Search~(ANNS) is a fundamental problem in computer science, particularly within domains that handle high-dimensional data such as computer vision~\cite{ben2015approximate} and multimedia~\cite{ferhatosmanoglu2001approximate}. ANNS aims to find points in a dataset that are close to a query point, typically in high-dimensional spaces, where searching the exact nearest neighbor is computationally prohibitive~\cite{liu2004investigation}.
One of the early approaches, KD-tree~\cite{friedman1977algorithm}, performs well in low-dimensional spaces. But their efficiency decreases significantly as dimensionality increases, a phenomenon known as the "curse of dimensionality". To address the limitation, more sophisticated tree-based structures like Randomized KD-trees~\cite{silpa2008optimised} attempt to partition the data space more effectively and offer improved performance in higher-dimensional spaces.
As research progressed, hashing-based methods emerged as a powerful alternative for ANNS, with Locality-Sensitive Hashing~(LSH) being a prominent technique to hash input items by mapping similar items to the same "buckets" with high probability~\cite{indyk1998approximate}.
More recently, graph-based methods have gained popularity for their effectiveness in handling high-dimensional ANNS~\cite{wang2021comprehensive}. The idea is to construct a graph where nodes represent the dataset points, and edges connect points close to each other. Navigating this graph allows for efficient ANN searches.
Notably, the Navigable Small World~(NSW)~\cite{malkov2014approximate} and its improved variant, Hierarchical Navigable Small World (HNSW)~\cite{malkov2018efficient}, have shown remarkable performance, significantly outperforming previous methods on large-scale datasets.
Besides, instead of Delaunay graphs, other graph-based approaches choose K-nearest neighbor graph~\cite{fu2016efanna} and relative neighborhood graph~\cite{jayaram2019diskann} for index construction.
In industrial applications, SL2G~\cite{tan2020fast} first leverages DNN as the relevance metric for graph-based ANNS in real-world recommendations.
NANN~\cite{chen2022approximate} extends SL2G by improving the DNN-based metric for better accuracy and adaptivity and working ANNS with controllable costs for industrial requirements.
In this work, we propose a parallel retrieval algorithm for web-scale recommendations with wider search space and better concurrent computational efficiency.

\section{Conclusion}

This work investigates the retrieval stage optimization for the web-scale recommendation. We propose a GMP-GR framework to enhance the retrieval accuracy and efficiency.
First, we devise multi-relational user-item relevance metric learning to improve user-item multi-relational understanding and graph search efficiency in a multi-objective way.
Second, we design a tailored parallel graph-based ANNS algorithm to distributedly broaden search space and parallelize irregular neural computations on GPUs.
Besides, we employ system optimization strategies to improve online recommendation performance.
We conduct extensive experiments on offline datasets and online services to demonstrate the outperformance of GMP-GR.
We have implemented GMP-GR in the recommender systems at Baidu, where it serves hundreds of millions of users and processes more than one hundred billion online requests per day.


\clearpage
\bibliographystyle{ACM-Reference-Format}
\bibliography{reference}

\clearpage
\appendix

\section{Appendix}

\subsection{Pseudo Code of HiPANNS Algorithm}\label{sec:appendix_pseudocode}

This section presents the pseudocodes of the Hierarchically Parallel Approximate Nearest Neighbor Search~(HiPANNS), including inter-candidate HiPANNS~(C-HiPANNS) and inter-query~(Q-HiPANNS).
In Algorithm~\ref{alg:c_hipanns}, inter-candidate HiPANNS~(C-HiPANNS) allows parallel search within the large-scale graph item base to broaden the global search space for more comprehensive retrieval.
In Algorithm~\ref{alg:q_hipanns}, inter-query HiPANNS~(Q-HiPANNS) adaptively aggregates irregular neural computations for more efficient GPU-accelerated retrieval against launch bound from extreme concurrency.

\begin{algorithm}
    \caption{Inter-candidate HiPANNS.}
    \label{alg:c_hipanns}
    \LinesNumbered
    \KwIn{Query user $u$, HNSW $G$, Neighbor hops $T$, heap size $K$, and number of HNSW layers $L$.}
    \KwOut{Top-$K$ retrieved items $V_u$.}
    \Fn{\CHIPANNS{$u$,$G$,$T$,$K$,$L$}}{
        Initialize a $K$-size max heap $\Phi_u$\;
        Initialize $K$ search seeds $S^{(0)}=\{v^{(0)}_{1},v^{(0)}_{2},\cdots,v^{(0)}_{K}\}$ and push them into $\Phi_u$\;
        \For{$l=1,\cdots,L$}{
            Initialize search seeds for $l$-th layer as $S^{(l)}=\Phi_u$\;
            \ForEach{$v^{(l)}_{i} \in S^{(l)}$ in parallel}{
                \For{$t=1,\cdots,T$ in batch}{
                    Retrieve $v^{(l)}_{i}$'s $t$-hop neighbor set $\{v^{(l)}_{i,t_j}\}$\;
                    Calculate the relevance $\delta_{u,v^{(l)}_{i,t_j}}$ by \QHIPANNS{$\{(u,v^{(l)}_{i,t_j})\}$}\;
                    Push $v^{(l)}_{i,t_j}$ into $\Phi_u$\;
                }
            }
        }
        \Return{$\Phi_u$}
    }
\end{algorithm}

\begin{algorithm}
    \caption{Inter-query HiPANNS}
    \label{alg:q_hipanns}
    \LinesNumbered
    \KwIn{GPU cluster $U$, $U$'s batch size $N_U$}
    \KwOut{Relevance values $\delta_{B_i}$}
    \Fn{\QHIPANNS{$B_i$}}{
        Initialize an empty $N_U$-size batch $B_U$ and current batch length $N^{\prime}_U=0$\;
        Initialize a queue $\Psi$\;
        Listen to a batch $B_i=\{(u_{i,1},v_{i,1}),\cdots,(u_{i,N_{B_i}},v_{i,N_{B_i}})\}$ at real-time and push it into $\Psi$\;
        \ForEach{$B_i \in \Psi$}{
            $\epsilon = N_U - N^{\prime}_U$\;
            \eIf{$N_{B_i} \leq \epsilon$}{
                Concentrate batch as $B_U = B_U \cap B_i$ and $N^{\prime}_U = N^{\prime}_U + N_{B_i}$\;
                Calculate $\delta_{B_i}$ on $U$\;
            }{
                Split $B_i$ as $B_i^{[1,\epsilon]}$ and $B_i^{[\epsilon+1,N_{B_i}]}$\;
                Concentrate batch as $B_U = B_U \cap B_i^{[1,\epsilon]}$\;
                Calculate $\delta_{B_i}$ on $U$\;
                Set $B_U = B_i^{[\epsilon+1,N_{B_i}]}$ and $N^{\prime}_U = B_i - \epsilon$\;
            }
        }
        Concentrate $\delta_{B_i}$ if $B_i$ is splited\;
        \Return{$\delta_{B_i}$}
    }
\end{algorithm}

\subsection{Implementation Details}\label{sec:appendix_implementation}

We construct HNSW by $4$ layers consistently for Video, MVideo, Explore, and Fresh datasets. Detailed statistics of layer-wise node numbers in HNSW structures are presented in Table~\ref{tab:nodes}.
Besides, we involve $35$ slots of features in user representation learning.
The neural metric has about $3$ million of learnable parameters.
For commercial online serving, we deploy our methods for $2$ services with $5$ servers, and activate $165$ and $119$ NVIDIA A30 Tensor Core GPUs, respectively. 
Detailed GPU equipment of servers is provided in Table~\ref{tab:gpu}.

\begin{table}
    \centering
    \caption{Number of nodes in HNSW layers for four datasets.}
    \label{tab:nodes}
    \begin{tabular}{c|cccc}
        \toprule
         & Video & MVideo & Explore & Fresh \\
        \midrule
        Layer 1 & 281 & 214 & 210 & 78 \\
        Layer 2 & 4,676 & 3,560 & 3,530 & 1,249 \\
        Layer 3 & 79,249 & 61,246 & 60,679 & 20,874 \\
        Layer 4 & 1,335,473 & 1,033,857 & 1,024,531 & 355,585 \\
        \bottomrule
    \end{tabular}
\end{table}

\begin{table}
    \centering
    \caption{GPU equipment of five servers for two services.}
    \label{tab:gpu}
    \begin{tabular}{c|ccccc}
        \toprule
         & Server 1 & Server 2 & Server 3 & Server 4 & Server 5 \\
        \midrule
        Service 1 & 28 & 24 & 36 & 40 & 37 \\
        Service 2 & 23 & 15 & 29 & 29 & 23 \\
        \bottomrule
    \end{tabular}
\end{table}

\clearpage

\subsection{Additional Efficiency Analysis}\label{sec:appendix_efficiency}

\begin{figure}[h]
    \centering
    \subfigure[MVideo]{\includegraphics[width=0.48\linewidth]{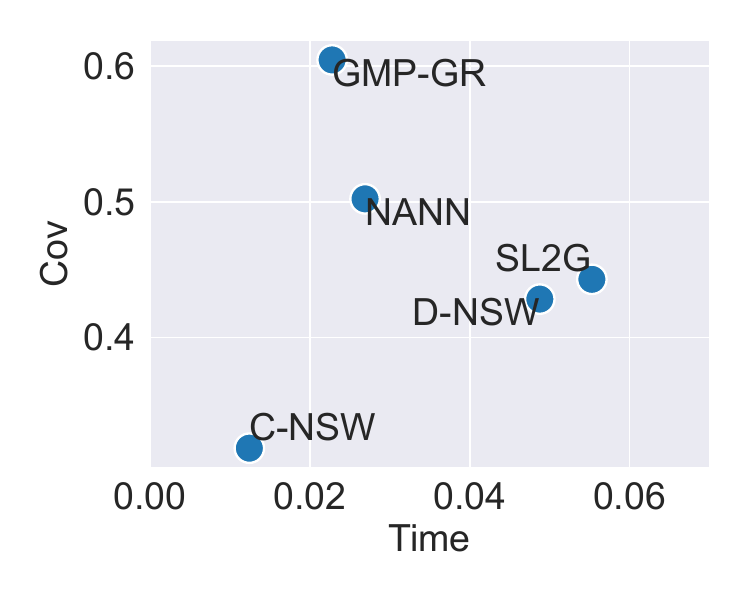}\label{fig:effciency_mvideo}}
    \subfigure[Fresh]{\includegraphics[width=0.48\linewidth]{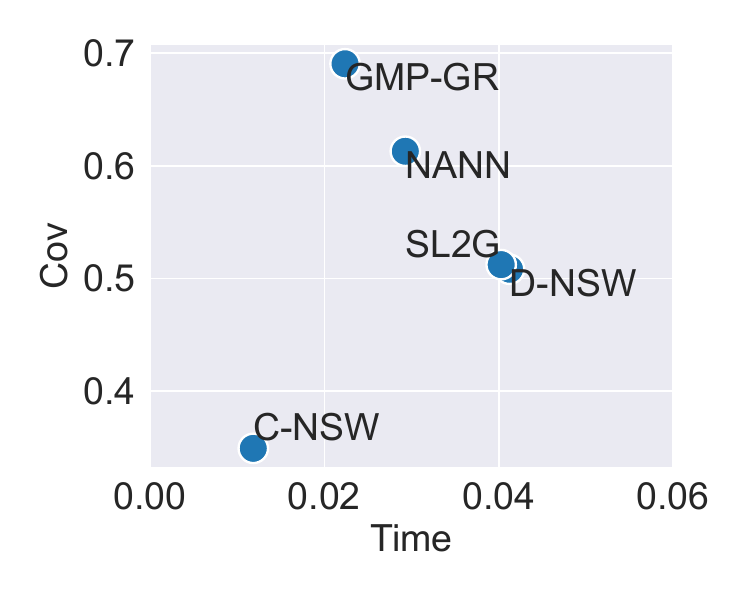}\label{fig:fig:effciency_fresh}}
    \caption{Analysis of correlations between retrieval performance and efficiency on four datasets. Methods close to the left~top~($\nwarrow$) have better retrieval performance, \ie Cov~($\uparrow$), with less time costs~($\leftarrow$).}
    \label{fig:appendix_efficiency}
\end{figure}

Here we analyze the efficiency in terms of inference time and search speed on MVideo and Fresh datasets by scattering results of five methods in Figure~\ref{fig:appendix_efficiency}.

\subsection{Additional Ablation Study}\label{sec:appendix_ablation}

\begin{figure}[h]
    \centering
    \subfigure[Rec@10]{\includegraphics[width=0.48\linewidth]{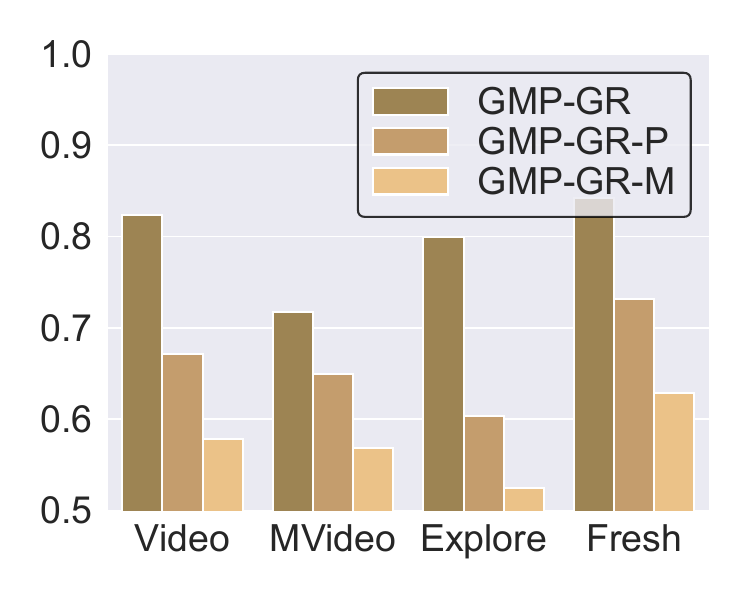}\label{fig:ablation_rec10}}
    \subfigure[Rec@100]{\includegraphics[width=0.48\linewidth]{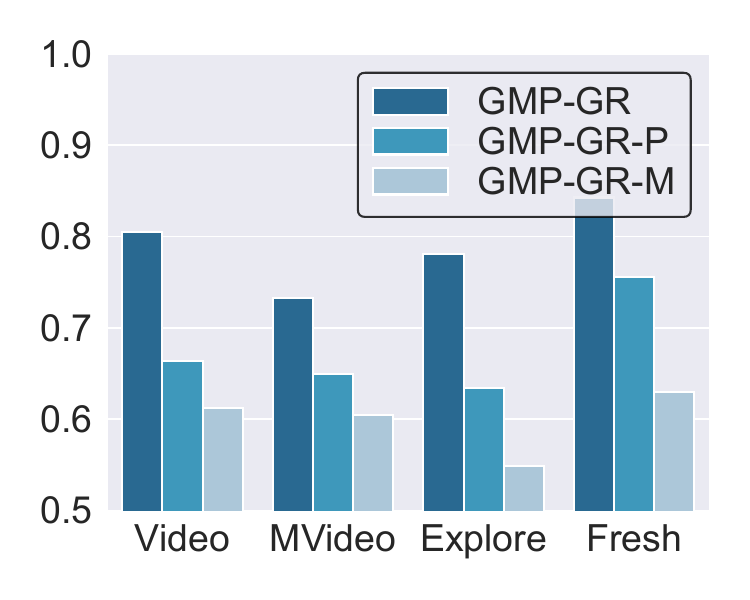}\label{fig:ablation_rec100}}
    \caption{Additional ablation study of GMP-GR on retrieval performance~(Rec@10 and Rec@100).}
    \label{fig:appendix_ablation}
\end{figure}

In this subsection, we conduct a further ablation study by comparing GMP-GR with GMP-GR-P and GMP-GR-M according to the evaluated results by Rec@10 and Rec@100 in Figure~\ref{fig:appendix_ablation}.

\end{document}